\documentclass{article}
\usepackage{spconf,amsmath,graphicx}
\usepackage[hidelinks]{hyperref}
\usepackage{booktabs}
\usepackage{tikz}
\usepackage{pgfplots}
\pgfplotsset{compat=1.3}
\usepackage{subcaption}
\usepackage{xcolor}
\usepackage{booktabs}
\usepackage{multirow}
\usepackage{cite}
\usepackage{amsfonts}

\newcommand{\Lc}{\mathcal{L}}

\newcommand{\Rb}{\mathbb{R}}

\newcommand{\gv}{\mathbf{g}}

\newcommand{\pv}{\mathbf{p}}

\newcommand{\xv}{\mathbf{x}}

\newcommand{\zv}{\mathbf{z}}

\newcommand{\Xv}{\mathbf{X}}
\newcommand{\Yv}{\mathbf{Y}}

\newcommand{\alphav     }{\boldsymbol \alpha     }

\title{I3D: Transformer architectures with input-dependent dynamic depth for speech recognition}
\name{Yifan Peng$^1$, Jaesong Lee$^2$, Shinji Watanabe$^1$}
\address{$^1$Carnegie Mellon University \quad $^2$NAVER Corporation}
\begin{document}
\ninept
\maketitle
\begin{abstract}
Transformer-based end-to-end speech recognition has achieved great success. However, the large footprint and computational overhead make it difficult to deploy these models in some real-world applications. Model compression techniques can reduce the model size and speed up inference, but the compressed model has a fixed architecture which might be suboptimal. We propose a novel Transformer encoder with \textbf{I}nput-\textbf{D}ependent \textbf{D}ynamic \textbf{D}epth (I3D) to achieve strong performance-efficiency trade-offs. With a similar number of layers at inference time, I3D-based models outperform the vanilla Transformer and the static pruned model via iterative layer pruning. We also present interesting analysis on the gate probabilities and the input-dependency, which helps us better understand deep encoders.
\end{abstract}
\begin{keywords}
Dynamic depth, transformer, speech recognition
\end{keywords}

\section{Introduction}
\label{sec:intro}

Recently, end-to-end automatic speech recognition (ASR) has gained popularity. Typical frameworks include Connectionist Temporal Classification (CTC)~\cite{ctc}, Attention-based Encoder-Decoder (AED)~\cite{aed-1, aed-2, las}, and Recurrent Neural Network Transducer (RNN-T)~\cite{rnnt}. Many types of networks can be used as encoders in these frameworks, such as Convolutional Neural Networks (CNNs), RNNs, Transformers~\cite{transformer} and their combinations~\cite{conformer, branchformer, e-branchformer}. Transformers have achieved great success in various benchmarks~\cite{karita2019comparative}. However, they usually contain many cascaded blocks and thus have high computation, which hinders deployment in some real-world applications with limited resource. To reduce computation and speed up inference, researchers have investigated different approaches.

A popular method is to compress a large pre-trained model using distillation~\cite{hinton2015distilling, distilhubert, lighthubert}, pruning~\cite{prune-rnn, tan2021compressing, lai2021parp}, and quantization~\cite{tan2021compressing}. However, the compressed model has a fixed architecture for all types of inputs, which might be suboptimal. For example, this fixed model may be too expensive for very easy utterances but insufficient for difficult ones. To better trade off performance and computation, prior studies have explored dynamic models~\cite{dynamic-survey-tpami}, which can adapt their architectures to different inputs. Dynamic models have shown to be effective in computer vision~\cite{bengio2015conditional, conv-aig, wang2018skipnet,wu2018blockdrop, shen2020fractionalskipping, li2021dynamicslimmable}, which are mainly based on CNNs. For speech processing, \cite{macoskey2021bifocal} trains two RNN encoders of different sizes and dynamically switches between them guided by keyword spotting. \cite{shi2021dynamic-encoder-transducer} proposes a dynamic encoder transducer based on layer dropout and collaborative learning. \cite{weninger2021dualencoder} adopts two RNN encoders to tackle close-talk and far-talk speech. \cite{macoskey2021amortized} also designs two RNN encoders that are compressed to different degrees and switches between them on a frame-by-frame basis. \cite{xie22interspeech-streaming} extends this idea to Transformer-transducers and considers more flexible subnetworks, but it continues to focus on streaming ASR and the architecture is still determined on a frame-by-frame basis, which requires a special design for the fined-grained key and query operations in self-attention. 
For nonstreaming (or chunk-based streaming) ASR, the frame-level prediction may be expensive and suboptimal, as it only captures frame-level local features.

We propose a Transformer encoder with \textbf{I}nput-\textbf{D}ependent \textbf{D}ynamic \textbf{D}epth (I3D) for end-to-end ASR. Instead of using carefully designed fine-grained operations within submodules like attention, I3D predicts whether to skip an entire self-attention block or an entire feed-forward network through a series of local gate predictors or a single global gate predictor. The prediction is made at the utterance level (or chunk level if extended to streaming cases), which is easier to implement and reduces additional cost. It also captures global statistics of the input. As analyzed in Sec.~\ref{subsec:analysis-input-dependency}, the length of an utterance affects the inference architecture. Some blocks may be useful for longer inputs.
Results show that I3D models consistently outperform Transformers trained from scratch and the static pruned models via iterative layer pruning~\cite{lee2021layer}, when using a similar number of layers for inference. We also perform interesting analysis on predicted gate probabilities and input-dependency, which helps us better understand the behavior of deep encoders.

\begin{figure*}[t]
     \centering
     \begin{subfigure}[b]{0.25\textwidth}
         \centering
         \includegraphics[width=0.82\textwidth]{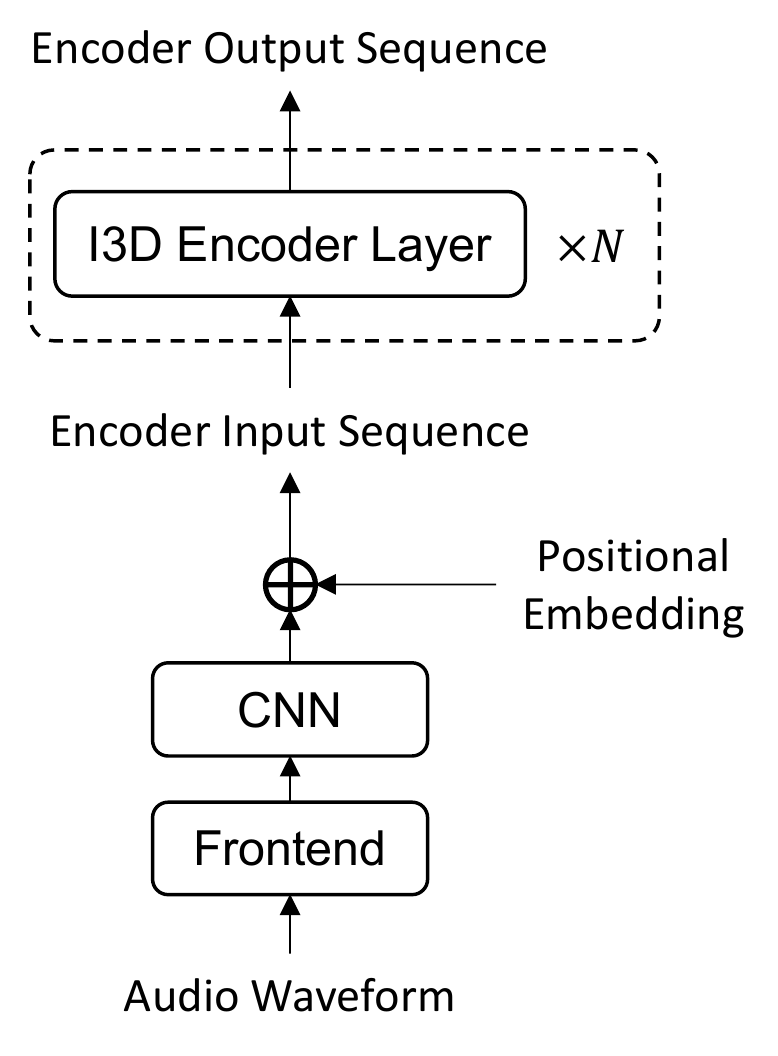}
         \vskip -0.1in
         \caption{Overall encoder architecture.}
         \label{fig:i3d-overall}
     \end{subfigure}
     \hfill
     \begin{subfigure}[b]{0.34\textwidth}
         \centering
         \includegraphics[width=0.64\textwidth]{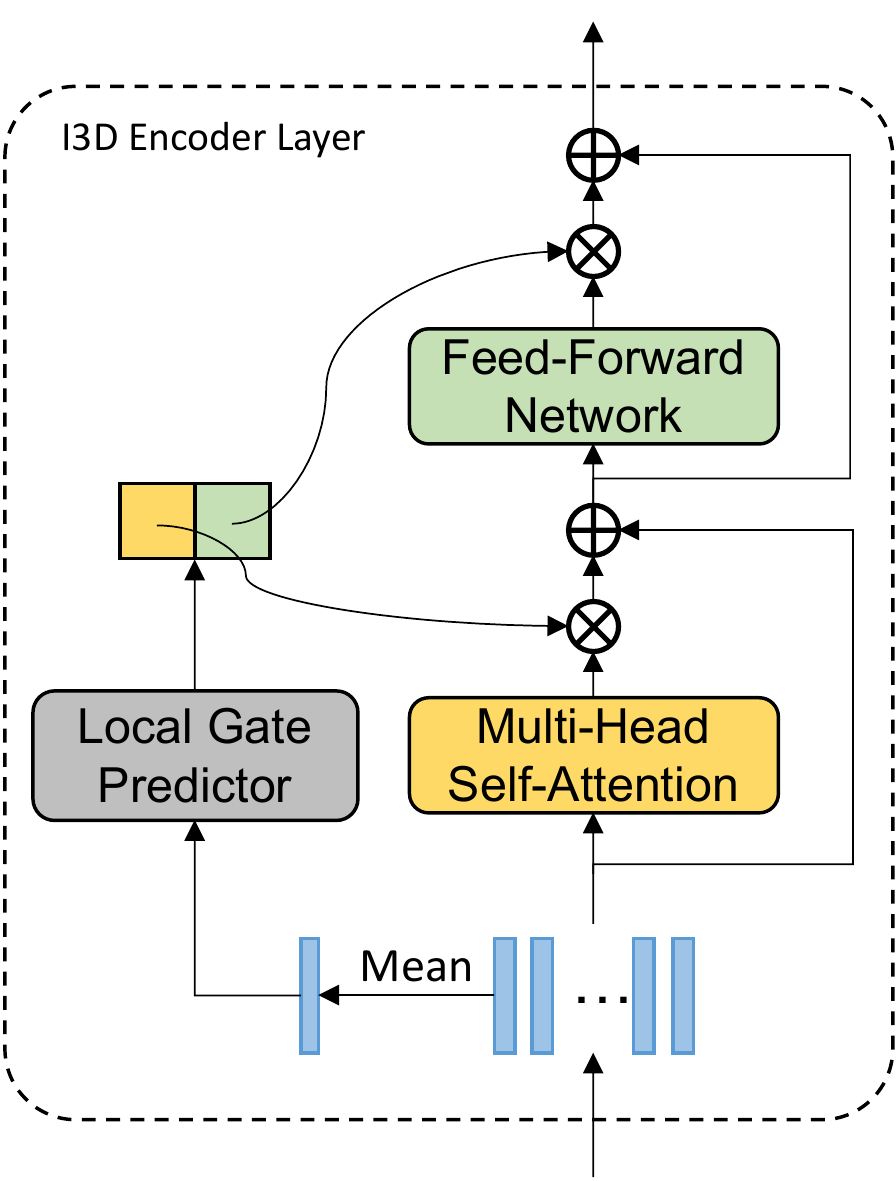}
         \vskip -0.1in
         \caption{I3D encoder layer with a local gate predictor.}
         \label{fig:i3d-local}
     \end{subfigure}
     \hfill
     \begin{subfigure}[b]{0.39\textwidth}
         \centering
         \includegraphics[width=0.75\textwidth]{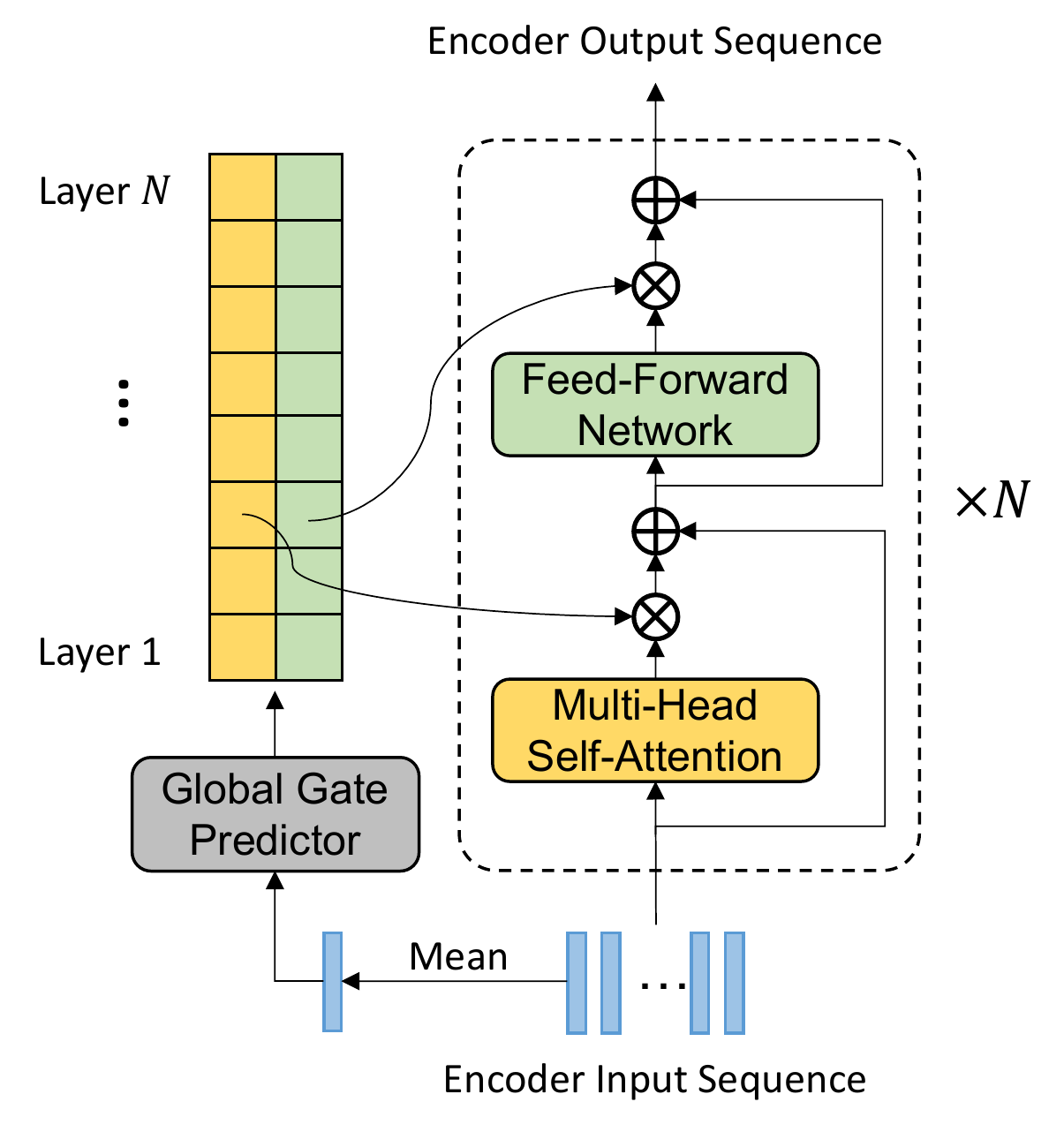}
         \vskip -0.1in
         \caption{I3D encoder with a global gate predictor.}
         \label{fig:i3d-global}
     \end{subfigure}
        \vskip -0.13in
        \caption{Architectures of our proposed I3D encoders.}
        \label{fig:model-architectures}
        \vskip -0.2in
\end{figure*}

\section{Method}
\label{sec:method}

\subsection{Transformer encoder}
\label{subsec:transformer-encoders}

A Transformer~\cite{transformer} encoder layer contains a multi-head self-attention (MHA) module and a feed-forward network (FFN), which are combined sequentially. The function of the $l$-th layer is as follows:
\begin{align}
    \Yv^{(l)} &= \Xv^{(l-1)} + \text{MHA}^{(l)}(\Xv^{(l-1)}), \label{eq:mha} \\
    \Xv^{(l)} &= \Yv^{(l)} + \text{FFN}^{(l)}(\Yv^{(l)}), \label{eq:ffn}
\end{align}
where $\Xv^{(l)}$ is the output of the $l$-th Transformer layer and $\Xv^{(l-1)}$ is thus the input to the $l$-th layer. $\Yv^{(l)}$ is the output of the MHA at the $l$-th layer, which is also the input of the FFN at the $l$-th layer. These sequences all have a length of $T$ and a feature size of $d$.

\subsection{Overall architecture of I3D encoders}
\label{subsec:overall-architecture}

Fig.~\ref{fig:i3d-overall} shows the overall architecture of I3D encoders. A waveform is first converted to a feature sequence by a frontend, and further processed and downsampled by a CNN, after which positional embeddings are added. Later, the sequence is processed by a stack of $N$ I3D encoder layers to produce high-level features. This overall design follows that of Transformer. However, Transformer always uses a fixed architecture regardless of the input. Our I3D selects different combinations of MHA and FFN depending on the input utterance. 
To determine whether a module should be executed or skipped, a \textit{binary gate} is introduced for each MHA or FFN module. The function of the $l$-th layer (see Eqs.~\eqref{eq:mha} and \eqref{eq:ffn} for the vanilla Transformer) now becomes:
\begin{align}
    \Yv^{(l)} &= \Xv^{(l-1)} + g^{(l)}_\text{MHA} \cdot \text{MHA}^{(l)}(\Xv^{(l-1)}), \label{eq:mha-gated} \\
    \Xv^{(l)} &= \Yv^{(l)} + g^{(l)}_\text{FFN} \cdot \text{FFN}^{(l)}(\Yv^{(l)}), \label{eq:ffn-gated}
\end{align}
where $g^{(l)}_\text{MHA}, g^{(l)}_\text{FFN} \in \{0, 1\}$ are input-dependent gates. If a gate is predicted to be 0, then the corresponding module will be skipped, which effectively reduces computation. The total training loss is:
\begin{align}
    \Lc_\text{total} &= \Lc_\text{ASR} + \lambda \cdot \Lc_\text{utility}, \label{eq:total-loss}\\
    \Lc_\text{utility} &= \frac{1}{2N} \sum_{l=1}^N \left( g^{(l)}_\text{MHA} + g^{(l)}_\text{FFN} \right), \label{eq:utility-loss}
\end{align}
where $\Lc_\text{ASR}$ is the standard ASR loss and $\Lc_\text{utility}$ is a regularization loss measuring the utility rate of all MHA and FFN modules. $\lambda > 0$ is a hyper-parameter to trade off the recognition accuracy and computational cost.~\footnote{Our method can be extended to consider different costs of MHA and FFN. In Eq. (6), we can use a weighted average of the two types of gates, where the weights depend on their computational costs. Then, the training will minimize the overall computation instead of simply the number of layers.}
Note that the utility loss in Eq.~\eqref{eq:utility-loss} is defined for an individual utterance so the utterance index is omitted. In practice, a mini-batch is used and the loss is averaged over utterances.

A major issue with this training objective is that binary gates are not differentiable. To solve this problem, we apply the Gumbel-Softmax~\cite{gumbel-softmax, concrete-distribution} trick, which allows drawing hard (or soft) samples from a discrete distribution. Consider a discrete random variable $Z$ with probabilities $P(Z=k) \propto \alpha_k$ for any $k =1,\ldots,K$. To draw a sample from this distribution, we can first draw $K$ i.i.d. samples $\{g_k\}_{k=1}^K$ from the standard Gumbel distribution and then select the index with the largest perturbed log probability:
\begin{align}
    z = \operatorname*{arg\,max}_{k\in\{1,\ldots,K\}} \log \alpha_k + g_k.
\end{align}
The argmax is not differentiable, which can be relaxed to the softmax. It is known that any sample from a discrete distribution can be denoted as a one-hot vector, where the index of the only non-zero entry is the desired sample. With this vector-based notation, we can draw a soft sample as follows:
\begin{align}
    \zv = \text{softmax}\left((\log \alphav + \gv)/\tau \right), \label{eq:gumbel-softmax}
\end{align}
where $\alphav = (\alpha_1, \ldots, \alpha_K)$, $\gv = (g_1, \ldots, g_K)$, and $\tau$ is a temperature constant. Eq.~\eqref{eq:gumbel-softmax} is an approximation of the original sampling process, but it is differentiable w.r.t. $\alphav$ and thus suitable for gradient-based optimization. As $\tau \rightarrow 0$, the approximation becomes closer to the discrete version. We use $\tau=1$ in our experiments.

For the $l$-th MHA, a discrete probability distribution $\pv^{(l)}_\text{MHA} \in \Rb^2$ over two possible gate values (0 and 1) is predicted, where 0 means skipping this module and 1 means executing it. Then, a soft sample is drawn from this discrete distribution using Eq.~\eqref{eq:gumbel-softmax}, which is used as the gate in Eq.~\eqref{eq:mha-gated} during training. Similarly, for the $l$-th FFN, a distribution $\pv^{(l)}_\text{FFN} \in \Rb^2$ is predicted, from which a soft gate is drawn and used in Eq.~\eqref{eq:ffn-gated}. The gate distributions are generated by a \emph{gate predictor} based on the input features, as defined in Sec.~\ref{subsec:gate-predictors}.

\subsection{Local and global gate predictors}
\label{subsec:gate-predictors}

We propose two types of gate predictors, namely the \emph{local gate predictor} and \emph{global gate predictor}. We employ a multi-layer perceptron (MLP) with a single hidden layer of size 32 in all experiments, which has little computational overhead.

The \emph{local gate predictor} (LocalGP or LGP) is associated with a specific I3D encoder layer, as illustrated in Fig.~\ref{fig:i3d-local}. Every layer has its own gate predictor whose parameters are independent. Consider the $l$-th encoder layer with an input sequence $\Xv^{(l-1)} \in \Rb^{T\times d}$. This sequence is first converted to a $d$-dimensional vector $\xv^{(l-1)} \in \Rb^{d}$ through average pooling along the time dimension. Then, this pooled vector is transformed to two 2-dimensional probability vectors for the MHA gate and the FFN gate, respectively:
\begin{align}
    \pv^{(l)}_\text{MHA}, \pv^{(l)}_\text{FFN} = \text{LGP}^{(l)}(\xv^{(l-1)}),
\end{align}
where $\pv^{(l)}_\text{MHA}, \pv^{(l)}_\text{FFN} \in \Rb^2$ are introduced in Sec.~\ref{subsec:overall-architecture}, and $\text{LGP}^{(l)}$ is the local gate predictor at the $l$-th layer. With this formulation, the decision of executing or skipping any MHA or FFN module depends on the input to the current layer, which further depends on the decision made at the previous layer. Hence, the decisions are made sequentially from lower to upper layers. During inference, a fixed threshold $\beta \in [0,1]$ is utilized to produce a binary gate for every module:
\begin{align}
    g_\text{MHA}^{(l)} &= 1 ~~\text{if}~~ (\pv^{(l)}_\text{MHA})_1 > \beta ~~\text{else}~~ 0, \label{eq:mha-threshold}\\
    g_\text{FFN}^{(l)} &= 1 ~~\text{if}~~ (\pv^{(l)}_\text{FFN})_1 > \beta ~~\text{else}~~ 0, \label{eq:ffn-threshold}
\end{align}
where $(\pv^{(l)}_\text{MHA})_1$ is the probability of executing the MHA and $(\pv^{(l)}_\text{FFN})_1$ is the probability of executing the FFN. We use $\beta=0.5$ by default, but it is also possible to adjust the inference cost by changing $\beta$.

The \emph{global gate predictor} (GlobalGP or GGP), on the other hand, is defined for an entire I3D encoder, as shown in Fig.~\ref{fig:i3d-global}. It predicts the gate distributions for all layers based on the encoder's input, which is also the input to the first layer: $\Xv = \Xv^{(0)} \in \Rb^{T \times d}$. In particular, the sequence is transformed to a single vector $\xv = \xv^{(0)} \in \Rb^d$ by average pooling. Then, it is mapped to two sets of probability distributions for all $N$ MHA and FFN gates, respectively:
\begin{align}
    \{\pv^{(l)}_\text{MHA}\}_{l=1}^N, \{\pv^{(l)}_\text{FFN}\}_{l=1}^N = \text{GGP}(\xv),
\end{align}
where $\pv^{(l)}_\text{MHA}, \pv^{(l)}_\text{FFN} \in \Rb^2$ are the gate probability distributions at the $l$-th layer, and the I3D encoder has $N$ layers in total. Here, the decisions of executing or skipping modules are made immediately after seeing the encoder's input, which has lower computational overhead than LocalGP and allows for more flexible control over the inference architecture. During inference, we can still use a fixed threshold $\beta \in [0,1]$ to generate binary gates as in Eqs.~\eqref{eq:mha-threshold} and \eqref{eq:ffn-threshold}. 

\section{Experiments}
\label{sec:exp}

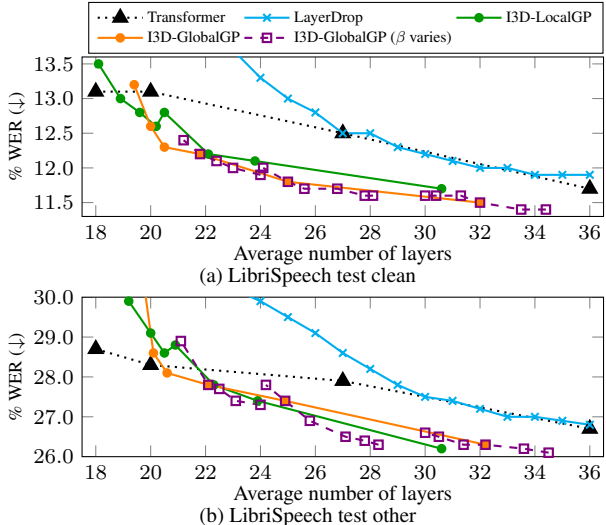
\begin{figure}[t]
     \centering
     \begin{subfigure}[b]{0.99\linewidth}
         \centering
         \begin{tikzpicture}
	\begin{axis}[
		xlabel=Average number of layers,
		ylabel=\% WER ($\downarrow$),
		xtick={0,2,...,100},
            ytick={11.5,12.0,12.5,...,14.0},
    	xmin=17.5,
            xmax=36.5,
            ymin=11.3,
            ymax=13.6,
            ylabel shift = -3pt,
            xlabel shift = -3.5pt,
            label style={font=\footnotesize},
            ylabel style={font=\scriptsize},
            ticklabel style={font=\footnotesize},
            yticklabel style={/pgf/number format/.cd,fixed,fixed zerofill,precision=1},
            width=\linewidth,
            height=0.435\linewidth,
	    every axis plot/.append style={thick},
            legend cell align={left},
            legend columns=3,
            legend style={at={(1,1.02)},anchor=south east,nodes={scale=0.66, transform shape}}
		]

        \addplot[color=black, dotted, mark=triangle*, mark options={scale=1.5, solid, fill=black}] coordinates {
(18.0, 13.1)
(20.0, 13.1)
(27.0, 12.5)
(36.0, 11.7)
	};
        \addlegendentry{Transformer};

\addplot[color=cyan, mark=x, solid, mark options={scale=1, solid, fill=cyan}] coordinates {
(36, 11.9)
(35, 11.9)
(34, 11.9)
(33, 12.0)
(32, 12.0)
(31, 12.1)
(30, 12.2)
(29, 12.3)
(28, 12.5)
(27, 12.5)
(26, 12.8)
(25, 13.0)
(24, 13.3)
(23, 13.7)
(22, 14.1)
(21, 14.6)
(20, 15.2)
(19, 15.6)
(18, 16.5)
};
        \addlegendentry{LayerDrop};

	\addplot[color=green!60!black, solid, mark=*, mark options={scale=0.7, solid}] coordinates {
(30.6, 11.7)
(23.8, 12.1)
(22.1, 12.2)
(20.5, 12.8)
(20.2, 12.6)
(19.6, 12.8)
(18.9, 13.0)
(18.1, 13.5)
	};
        \addlegendentry{I3D-LocalGP};
        
	\addplot[color=orange, mark=*, solid, mark options={scale=0.7, solid}] coordinates {
(32.0, 11.5)
(25.0, 11.8)
(21.8, 12.2)
(20.5, 12.3)
(20.0, 12.6)
(19.4, 13.2)
	};
        \addlegendentry{I3D-GlobalGP};

        \addplot[color=violet, mark=square, dashed, mark options={scale=0.8, solid}] coordinates {
(34.4, 11.4)
(33.5, 11.4)
(32.0, 11.5)
(31.3, 11.6)
(30.4, 11.6)
(30.0, 11.6)
	};
        \addlegendentry{I3D-GlobalGP ($\beta$ varies)};

        \addplot[color=violet, mark=square, dashed, mark options={scale=0.8, solid}] coordinates {
(28.1, 11.6)
(27.8, 11.6)
(26.8, 11.7)
(25.6, 11.7)
(25.0, 11.8)
(24.1, 12.0)

	};

        \addplot[color=violet, mark=square, dashed, mark options={scale=0.8, solid}] coordinates {
(24.0, 11.9)
(23.0, 12.0)
(22.4, 12.1)
(21.8, 12.2)
(21.2, 12.4)
	};

	\end{axis}
\end{tikzpicture}
         \vskip -0.12in
         \caption{LibriSpeech test clean}
         \label{fig:ls-ctc-test-clean}
     \end{subfigure}
     \vskip -0.05in
     
     \begin{subfigure}[b]{0.99\linewidth}
         \centering
         \begin{tikzpicture}
	\begin{axis}[
		xlabel=Average number of layers,
		ylabel=\% WER ($\downarrow$),
		xtick={0,2,...,100},
            ytick={0.0,1.0,...,100.0},
    	xmin=17.5,
            xmax=36.5,
            ymin=26,
            ymax=30,
            ylabel shift = -3pt,
            xlabel shift = -3.5pt,
            label style={font=\footnotesize},
            ylabel style={font=\scriptsize},
            ticklabel style={font=\footnotesize},
            yticklabel style={/pgf/number format/.cd,fixed,fixed zerofill,precision=1},
            width=\linewidth,
            height=0.435\linewidth,
	    every axis plot/.append style={thick},
            legend cell align={left},
            legend columns=1,
            legend style={at={(1,1)},anchor=north east,nodes={scale=0.65, transform shape}}
		]

        \addplot[color=black, dotted, mark=triangle*, mark options={scale=1.5, solid, fill=black}] coordinates {
(18.0, 28.7)
(20.0, 28.3)
(27.0, 27.9)
(36.0, 26.7)
	};

\addplot[color=cyan, mark=x, solid, mark options={scale=1, solid, fill=cyan}] coordinates {
(36, 26.8)
(35, 26.9)
(34, 27.0)
(33, 27.0)
(32, 27.2)
(31, 27.4)
(30, 27.5)
(29, 27.8)
(28, 28.2)
(27, 28.6)
(26, 29.1)
(25, 29.5)
(24, 29.9)
(23, 30.2)
(22, 31.0)
(21, 31.7)
(20, 32.6)
(19, 33.6)
(18, 34.3)
};

	\addplot[color=green!60!black, solid, mark=*, mark options={scale=0.7, solid}] coordinates {
(30.6, 26.2)
(23.9, 27.4)
(22.3, 27.8)
(20.9, 28.8)
(20.5, 28.6)
(20.0, 29.1)
(19.2, 29.9)
(18.4, 30.9)
	};
        
	\addplot[color=orange, mark=*, solid, mark options={scale=0.7, solid}] coordinates {
(32.2, 26.3)
(24.9, 27.4)
(22.1, 27.8)
(20.6, 28.1)
(20.1, 28.6)
(19.5, 31.6)
	};

        \addplot[color=violet, mark=square, dashed, mark options={scale=0.8, solid}] coordinates {
(34.5, 26.1)
(33.6, 26.2)
(32.2, 26.3)
(31.4, 26.3)
(30.5, 26.5)
(30.0, 26.6)
	};

        \addplot[color=violet, mark=square, dashed, mark options={scale=0.8, solid}] coordinates {
(28.3, 26.3)
(27.8, 26.4)
(27.1, 26.5)
(25.8, 26.9)
(24.9, 27.4)
(24.2, 27.8)
	};

        \addplot[color=violet, mark=square, dashed, mark options={scale=0.8, solid}] coordinates {
(24.0, 27.3)
(23.1, 27.4)
(22.5, 27.7)
(22.1, 27.8)
(21.1, 28.9)
	};

	\end{axis}
\end{tikzpicture}
         \vskip -0.12in
         \caption{LibriSpeech test other}
         \label{fig:ls-ctc-testother}
     \end{subfigure}
     \vskip -0.15in
        \caption{Word error rates (\%) of \textbf{CTC}-based models vs. average number of layers used for inference on \textbf{LibriSpeech} test sets. $\beta$ is the threshold for generating binary gates as defined in Eqs.~\eqref{eq:mha-threshold}~\eqref{eq:ffn-threshold}.}
        \label{fig:ls-ctc}
        \vskip -0.15in
\end{figure}

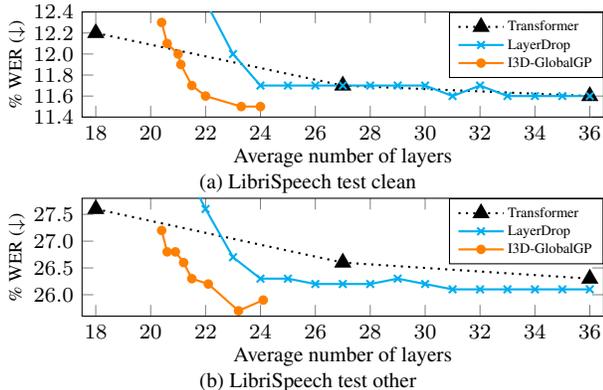
\begin{figure}[t]
     \centering
     \begin{subfigure}[b]{0.99\linewidth}
         \centering
         \begin{tikzpicture}
	\begin{axis}[
		xlabel=Average number of layers,
		ylabel=\% WER ($\downarrow$),
		xtick={0,2,...,100},
    	xmin=17.5,
            xmax=36.5,
            ymin=11.4,
            ymax=12.4,
            ylabel shift = -3pt,
            xlabel shift = -3pt,
            label style={font=\footnotesize},
            ylabel style={font=\scriptsize},
            ticklabel style={font=\footnotesize},
            yticklabel style={/pgf/number format/.cd,fixed,fixed zerofill,precision=1},
            width=\linewidth,
            height=0.35\linewidth,
	    every axis plot/.append style={thick},
            legend cell align={left},
            legend columns=1,
            legend style={at={(1,1)},anchor=north east,nodes={scale=0.6, transform shape}}
		]

        \addplot[color=black, dotted, mark=triangle*, mark options={scale=1.5, solid, fill=black}] coordinates {
(18.0, 12.2)
(27.0, 11.7)
(36.0, 11.6)
	};
        \addlegendentry{Transformer};

\addplot[color=cyan, mark=x, solid, mark options={scale=1, solid, fill=cyan}] coordinates {
(36, 11.6)
(35, 11.6)
(34, 11.6)
(33, 11.6)
(32, 11.7)
(31, 11.6)
(30, 11.7)
(29, 11.7)
(28, 11.7)
(27, 11.7)
(26, 11.7)
(25, 11.7)
(24, 11.7)
(23, 12.0)
(22, 12.5)
(21, 13.1)
(20, 13.6)
(19, 14.4)
(18, 15.6)

	};
        \addlegendentry{LayerDrop};
        
	\addplot[color=orange, mark=*, solid, mark options={scale=0.7, solid}] coordinates {
(24.0, 11.5)
(23.3, 11.5)
(22.0, 11.6)
(21.5, 11.7)
(21.1, 11.9)
(21.0, 12.0)
(20.6, 12.1)
(20.4, 12.3)
	};
        \addlegendentry{I3D-GlobalGP};

	\end{axis}
\end{tikzpicture}
         \vskip -0.1in
         \caption{LibriSpeech test clean}
         \label{fig:ls-interctc-test-clean}
     \end{subfigure}
     
     \begin{subfigure}[b]{0.99\linewidth}
         \centering
         \begin{tikzpicture}
	\begin{axis}[
		xlabel=Average number of layers,
		ylabel=\% WER ($\downarrow$),
		xtick={0,2,...,100},
    	xmin=17.5,
            xmax=36.5,
            ymin=25.6,
            ymax=27.8,
            ylabel shift = -3pt,
            xlabel shift = -3pt,
            label style={font=\footnotesize},
            ylabel style={font=\scriptsize},
            ticklabel style={font=\footnotesize},
            yticklabel style={/pgf/number format/.cd,fixed,fixed zerofill,precision=1},
            width=\linewidth,
            height=0.37\linewidth,
	    every axis plot/.append style={thick},
            legend cell align={left},
            legend columns=1,
            legend style={at={(1,1)},anchor=north east,nodes={scale=0.6, transform shape}}
		]

        \addplot[color=black, dotted, mark=triangle*, mark options={scale=1.5, solid, fill=black}] coordinates {
(18.0, 27.6)
(27.0, 26.6)
(36.0, 26.3)
	};
        \addlegendentry{Transformer};

\addplot[color=cyan, mark=x, solid, mark options={scale=1, solid, fill=cyan}] coordinates {
(36, 26.1)
(35, 26.1)
(34, 26.1)
(33, 26.1)
(32, 26.1)
(31, 26.1)
(30, 26.2)
(29, 26.3)
(28, 26.2)
(27, 26.2)
(26, 26.2)
(25, 26.3)
(24, 26.3)
(23, 26.7)
(22, 27.6)
(21, 28.6)
(20, 30.0)
(19, 31.4)
(18, 32.7)
	};
        \addlegendentry{LayerDrop};
        
	\addplot[color=orange, mark=*, solid, mark options={scale=0.7, solid}] coordinates {
(24.1, 25.9)
(23.2, 25.7)
(22.1, 26.2)
(21.5, 26.3)
(21.2, 26.6)
(20.9, 26.8)
(20.6, 26.8)
(20.4, 27.2)
	};
        \addlegendentry{I3D-GlobalGP};

	\end{axis}
\end{tikzpicture}
         \vskip -0.1in
         \caption{LibriSpeech test other}
         \label{fig:ls-interctc-testother}
     \end{subfigure}
     \vskip -0.15in
        \caption{Word error rates (\%) of \textbf{InterCTC}-based models vs. average number of layers used for inference on \textbf{LibriSpeech} test sets.}
        \label{fig:ls-interctc}
        \vskip -0.15in
\end{figure}

\begin{figure}[t]
\centering
\resizebox {0.99\linewidth} {!} {
\begin{tikzpicture}
	\begin{axis}[
		xlabel=Average number of layers,
		ylabel=\% WER ($\downarrow$),
		xtick={0,2,...,100},
    	xmin=17.5,
            xmax=36.5,
            ymin=10,
            ymax=13.5,
            ylabel shift = -3pt,
            xlabel shift = -3pt,
            label style={font=\scriptsize},
            ticklabel style={font=\footnotesize},
            width=\linewidth,
            height=0.33\linewidth,
	    every axis plot/.append style={thick},
            legend cell align={left},
            legend columns=-1,
            legend style={at={(1,1)},anchor=north east,nodes={scale=0.58, transform shape}}
		]

        \addplot[color=black, dotted, mark=triangle*, mark options={scale=1.5, solid, fill=black}] coordinates {
(18.0, 12.7)
(27.0, 11.4)
(36.0, 10.9)
	};
        \addlegendentry{Transformer};
        
	\addplot[color=green!60!black, solid, mark=*, mark options={scale=0.7, solid}] coordinates {
(29.4, 10.6)
(24.3, 11.3)
(22.0, 11.4)
(21.5, 11.8)
(21.0, 12.0)
(20.2, 12.2)
	};
        \addlegendentry{I3D-LocalGP};
        
	\addplot[color=orange, mark=*, solid, mark options={scale=0.7, solid}] coordinates {
(32.9, 10.5)
(29.3, 10.6)
(25.3, 10.8)
(24.2, 11.2)
(22.9, 11.2)
(21.4, 12.4)
	};
        \addlegendentry{I3D-GlobalGP};

	\end{axis}
\end{tikzpicture}
} 
    \vskip -0.15in
  \caption{Word error rates (\%) of \textbf{CTC}-based models vs. average number of layers used for inference on the \textbf{Tedlium2} test set.}
  \label{fig:tedlium-ctc-test}
  \vskip -0.1in
\end{figure}
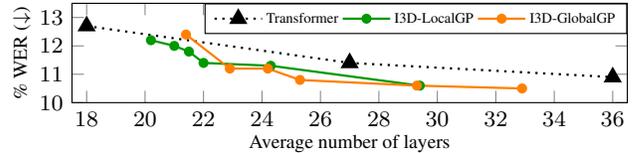
\begin{table}[t]
    \caption{Word error rates (\%) and average number of inference layers of \textbf{AED}-based models on \textbf{LibriSpeech 100h}.}
    \label{tab:ls100-aed}
    \vskip -0.1in
    \centering
    \resizebox {0.85\linewidth} {!} {
    \begin{tabular}{ccccc}
    \toprule
    \multirow{2}{*}{Model} & \multicolumn{2}{c}{dev clean} & \multicolumn{2}{c}{test clean}\\
    & Ave \#layers & WER ($\downarrow$) & Ave \#layers & WER ($\downarrow$) \\
    \midrule
    \multirow{2}{*}{Transformer} & 36 & 7.8 & 36 & 8.0 \\
     & 27 & 8.2 & 27 & 8.5 \\
    \midrule
    I3D-LGP-36 & 27.3 & 7.9 & 27.1 & 8.3\\
    I3D-GGP-36 & 27.2 & 7.8 & 27.1 & 8.2 \\
    \bottomrule
    \end{tabular}
}
\vskip -0.15in
\end{table}

\begin{figure}[t]
     \centering
     \begin{subfigure}[b]{\linewidth}
         \centering
         \resizebox {0.9\linewidth} {!} {
         \begin{tikzpicture}

\definecolor{darkgray176}{RGB}{176,176,176}
\definecolor{steelblue31119180}{RGB}{31,119,180}

\begin{axis}[
tick align=outside,
tick pos=left,
x grid style={darkgray176},
xlabel={Layer index},
xmin=0.5, xmax=37,
xtick style={color=black},
xtick={1,6,11,...,36},
y grid style={darkgray176},
ylabel={Probability},
ymin=-0.0500431648545307, ymax=1.04980744365575,
ytick style={color=black},
major tick length=2pt,
width=\linewidth,
height=0.33\linewidth,
ylabel shift = -3pt,
xlabel shift = -3pt,
label style={font=\footnotesize},
ticklabel style={font=\footnotesize},
]
\path [draw=steelblue31119180, semithick]
(axis cs:1,-1.21722898797626e-06)
--(axis cs:1,8.97713072609258e-06);

\path [draw=steelblue31119180, semithick]
(axis cs:2,-2.58555722141265e-05)
--(axis cs:2,9.88173871117142e-05);

\path [draw=steelblue31119180, semithick]
(axis cs:3,0.012243542524307)
--(axis cs:3,0.0352617315377002);

\path [draw=steelblue31119180, semithick]
(axis cs:4,0.701614267390356)
--(axis cs:4,0.908146701057105);

\path [draw=steelblue31119180, semithick]
(axis cs:5,0.000502799619334311)
--(axis cs:5,0.00265344135759359);

\path [draw=steelblue31119180, semithick]
(axis cs:6,-4.995537679069e-05)
--(axis cs:6,0.00314760235022492);

\path [draw=steelblue31119180, semithick]
(axis cs:7,0.000884380866033524)
--(axis cs:7,0.0062959010868226);

\path [draw=steelblue31119180, semithick]
(axis cs:8,-7.65848109970581e-07)
--(axis cs:8,1.23984941638476e-05);

\path [draw=steelblue31119180, semithick]
(axis cs:9,-9.01060855381881e-07)
--(axis cs:9,1.87613306115054e-05);

\path [draw=steelblue31119180, semithick]
(axis cs:10,-1.29609540287418e-06)
--(axis cs:10,1.28499035533559e-05);

\path [draw=steelblue31119180, semithick]
(axis cs:11,0.0155728906710551)
--(axis cs:11,0.0357622415835691);

\path [draw=steelblue31119180, semithick]
(axis cs:12,-1.12648407887397e-06)
--(axis cs:12,7.62680923041746e-06);

\path [draw=steelblue31119180, semithick]
(axis cs:13,0.288937931799343)
--(axis cs:13,0.497192629565656);

\path [draw=steelblue31119180, semithick]
(axis cs:14,0.91304290700657)
--(axis cs:14,0.965376807536908);

\path [draw=steelblue31119180, semithick]
(axis cs:15,0.0213143684723028)
--(axis cs:15,0.04785832460891);

\path [draw=steelblue31119180, semithick]
(axis cs:16,-1.15339590074327e-06)
--(axis cs:16,1.16580913770952e-05);

\path [draw=steelblue31119180, semithick]
(axis cs:17,0.849573343466243)
--(axis cs:17,0.911410231366285);

\path [draw=steelblue31119180, semithick]
(axis cs:18,0.0217110484343033)
--(axis cs:18,0.0718871299162044);

\path [draw=steelblue31119180, semithick]
(axis cs:19,0.92102743792186)
--(axis cs:19,0.962367536270434);

\path [draw=steelblue31119180, semithick]
(axis cs:20,-9.42304284539135e-07)
--(axis cs:20,6.27689672806027e-06);

\path [draw=steelblue31119180, semithick]
(axis cs:21,0.453805027810228)
--(axis cs:21,0.609101211063518);

\path [draw=steelblue31119180, semithick]
(axis cs:22,0.993731104926978)
--(axis cs:22,0.997580005611402);

\path [draw=steelblue31119180, semithick]
(axis cs:23,0.803074485436231)
--(axis cs:23,0.8866927292308);

\path [draw=steelblue31119180, semithick]
(axis cs:24,0.855623458693445)
--(axis cs:24,0.907989710213461);

\path [draw=steelblue31119180, semithick]
(axis cs:25,0.874984292519817)
--(axis cs:25,0.935616780239379);

\path [draw=steelblue31119180, semithick]
(axis cs:26,0.984580128547131)
--(axis cs:26,0.994540608809682);

\path [draw=steelblue31119180, semithick]
(axis cs:27,0.952156878745821)
--(axis cs:27,0.985371154825464);

\path [draw=steelblue31119180, semithick]
(axis cs:28,0.880371186688359)
--(axis cs:28,0.956036367343641);

\path [draw=steelblue31119180, semithick]
(axis cs:29,0.980761840489819)
--(axis cs:29,0.996433684200932);

\path [draw=steelblue31119180, semithick]
(axis cs:30,0.998716485531371)
--(axis cs:30,0.999814234178009);

\path [draw=steelblue31119180, semithick]
(axis cs:31,0.988299644588647)
--(axis cs:31,0.99800161082407);

\path [draw=steelblue31119180, semithick]
(axis cs:32,0.989822622170468)
--(axis cs:32,0.998595986865241);

\path [draw=steelblue31119180, semithick]
(axis cs:33,0.953129860306982)
--(axis cs:33,0.990152967848787);

\path [draw=steelblue31119180, semithick]
(axis cs:34,0.944121708362376)
--(axis cs:34,0.990913154707891);

\path [draw=steelblue31119180, semithick]
(axis cs:35,0.956760090411866)
--(axis cs:35,0.994910908894459);

\path [draw=steelblue31119180, semithick]
(axis cs:36,0.996772250608145)
--(axis cs:36,0.999794809147586);

\addplot [semithick, steelblue31119180, mark=-, mark size=2, mark options={solid}, only marks]
table {%
1 -1.21722898797626e-06
2 -2.58555722141265e-05
3 0.012243542524307
4 0.701614267390356
5 0.000502799619334311
6 -4.995537679069e-05
7 0.000884380866033524
8 -7.65848109970581e-07
9 -9.01060855381881e-07
10 -1.29609540287418e-06
11 0.0155728906710551
12 -1.12648407887397e-06
13 0.288937931799343
14 0.91304290700657
15 0.0213143684723028
16 -1.15339590074327e-06
17 0.849573343466243
18 0.0217110484343033
19 0.92102743792186
20 -9.42304284539135e-07
21 0.453805027810228
22 0.993731104926978
23 0.803074485436231
24 0.855623458693445
25 0.874984292519817
26 0.984580128547131
27 0.952156878745821
28 0.880371186688359
29 0.980761840489819
30 0.998716485531371
31 0.988299644588647
32 0.989822622170468
33 0.953129860306982
34 0.944121708362376
35 0.956760090411866
36 0.996772250608145
};
\addplot [semithick, steelblue31119180, mark=-, mark size=2, mark options={solid}, only marks]
table {%
1 8.97713072609258e-06
2 9.88173871117142e-05
3 0.0352617315377002
4 0.908146701057105
5 0.00265344135759359
6 0.00314760235022492
7 0.0062959010868226
8 1.23984941638476e-05
9 1.87613306115054e-05
10 1.28499035533559e-05
11 0.0357622415835691
12 7.62680923041746e-06
13 0.497192629565656
14 0.965376807536908
15 0.04785832460891
16 1.16580913770952e-05
17 0.911410231366285
18 0.0718871299162044
19 0.962367536270434
20 6.27689672806027e-06
21 0.609101211063518
22 0.997580005611402
23 0.8866927292308
24 0.907989710213461
25 0.935616780239379
26 0.994540608809682
27 0.985371154825464
28 0.956036367343641
29 0.996433684200932
30 0.999814234178009
31 0.99800161082407
32 0.998595986865241
33 0.990152967848787
34 0.990913154707891
35 0.994910908894459
36 0.999794809147586
};
\addplot [semithick, steelblue31119180, mark=*, mark size=1, mark options={solid}, only marks]
table {%
1 3.87995086905816e-06
2 3.64809074487938e-05
3 0.0237526370310036
4 0.804880484223731
5 0.00157812048846395
6 0.00154882348671711
7 0.00359014097642806
8 5.8163230269385e-06
9 8.93013487806176e-06
10 5.77690407524084e-06
11 0.0256675661273121
12 3.25016257577174e-06
13 0.3930652806825
14 0.939209857271739
15 0.0345863465406064
16 5.25234773817597e-06
17 0.880491787416264
18 0.0467990891752538
19 0.941697487096147
20 2.66729622176057e-06
21 0.531453119436873
22 0.99565555526919
23 0.844883607333515
24 0.881806584453453
25 0.905300536379598
26 0.989560368678407
27 0.968764016785643
28 0.918203777016
29 0.988597762345375
30 0.99926535985469
31 0.993150627706358
32 0.994209304517854
33 0.971641414077885
34 0.967517431535133
35 0.975835499653162
36 0.998283529877866
};
\end{axis}

\end{tikzpicture}
         }
         \vskip -0.12in
         \caption{MHA gate probabilities.}
         \label{fig:mha-gate-prob}
     \end{subfigure}

     \vskip -0.05in
     \begin{subfigure}[b]{\linewidth}
         \centering
         \resizebox {0.9\linewidth} {!} {
        \begin{tikzpicture}

\definecolor{darkgray176}{RGB}{176,176,176}
\definecolor{steelblue31119180}{RGB}{31,119,180}

\begin{axis}[
tick align=outside,
tick pos=left,
x grid style={darkgray176},
xlabel={Layer index},
xmin=0.5, xmax=37,
xtick style={color=black},
xtick={1,6,11,...,36},
y grid style={darkgray176},
ylabel={Probability},
ymin=-0.0504673897021493, ymax=1.050026171665,
ytick style={color=black},
major tick length=2pt,
width=\linewidth,
height=0.33\linewidth,
ylabel shift = -3pt,
xlabel shift = -3pt,
label style={font=\footnotesize},
ticklabel style={font=\footnotesize},
]
\path [draw=steelblue31119180, semithick]
(axis cs:1,0.999862696254946)
--(axis cs:1,1.0000037370574);

\path [draw=steelblue31119180, semithick]
(axis cs:2,-1.17940646205767e-06)
--(axis cs:2,1.14789987263998e-05);

\path [draw=steelblue31119180, semithick]
(axis cs:3,-0.000444955094551462)
--(axis cs:3,0.00064913240596353);

\path [draw=steelblue31119180, semithick]
(axis cs:4,0.149880880175691)
--(axis cs:4,0.345364907950286);

\path [draw=steelblue31119180, semithick]
(axis cs:5,0.0148539247429455)
--(axis cs:5,0.07314457421611);

\path [draw=steelblue31119180, semithick]
(axis cs:6,0.499177556359257)
--(axis cs:6,0.645458883771822);

\path [draw=steelblue31119180, semithick]
(axis cs:7,0.452171312470224)
--(axis cs:7,0.625949293071374);

\path [draw=steelblue31119180, semithick]
(axis cs:8,-1.66601383678422e-06)
--(axis cs:8,3.38226304607863e-05);

\path [draw=steelblue31119180, semithick]
(axis cs:9,-7.69951486654214e-07)
--(axis cs:9,7.45022603211534e-06);

\path [draw=steelblue31119180, semithick]
(axis cs:10,-3.65594625158578e-07)
--(axis cs:10,7.71478202587656e-06);

\path [draw=steelblue31119180, semithick]
(axis cs:11,-5.38016069689417e-07)
--(axis cs:11,1.00360531289019e-05);

\path [draw=steelblue31119180, semithick]
(axis cs:12,-6.50100828623347e-07)
--(axis cs:12,1.26173870491808e-05);

\path [draw=steelblue31119180, semithick]
(axis cs:13,-8.93660863582813e-07)
--(axis cs:13,1.24753944862336e-05);

\path [draw=steelblue31119180, semithick]
(axis cs:14,0.0755359300492911)
--(axis cs:14,0.10331609226345);

\path [draw=steelblue31119180, semithick]
(axis cs:15,0.110353076156504)
--(axis cs:15,0.155697896417795);

\path [draw=steelblue31119180, semithick]
(axis cs:16,0.317207567538609)
--(axis cs:16,0.362165645160042);

\path [draw=steelblue31119180, semithick]
(axis cs:17,0.287292123846153)
--(axis cs:17,0.357888287600546);

\path [draw=steelblue31119180, semithick]
(axis cs:18,0.0103364909681578)
--(axis cs:18,0.0191130795322117);

\path [draw=steelblue31119180, semithick]
(axis cs:19,0.61323344991875)
--(axis cs:19,0.660760820373833);

\path [draw=steelblue31119180, semithick]
(axis cs:20,0.70748251841905)
--(axis cs:20,0.754072902629127);

\path [draw=steelblue31119180, semithick]
(axis cs:21,0.545653313424864)
--(axis cs:21,0.633417729318876);

\path [draw=steelblue31119180, semithick]
(axis cs:22,0.74485186318899)
--(axis cs:22,0.800586261207122);

\path [draw=steelblue31119180, semithick]
(axis cs:23,0.786037264551979)
--(axis cs:23,0.848336852074331);

\path [draw=steelblue31119180, semithick]
(axis cs:24,0.846707330917601)
--(axis cs:24,0.910770201686911);

\path [draw=steelblue31119180, semithick]
(axis cs:25,0.908620153237746)
--(axis cs:25,0.952636265610178);

\path [draw=steelblue31119180, semithick]
(axis cs:26,0.939624328116863)
--(axis cs:26,0.972131526360618);

\path [draw=steelblue31119180, semithick]
(axis cs:27,0.958265356894392)
--(axis cs:27,0.985702041537859);

\path [draw=steelblue31119180, semithick]
(axis cs:28,0.992443957787352)
--(axis cs:28,0.998326739603532);

\path [draw=steelblue31119180, semithick]
(axis cs:29,0.981409930527113)
--(axis cs:29,0.993035794414477);

\path [draw=steelblue31119180, semithick]
(axis cs:30,0.974249181864862)
--(axis cs:30,0.993189633455799);

\path [draw=steelblue31119180, semithick]
(axis cs:31,0.974968447471581)
--(axis cs:31,0.992635409155749);

\path [draw=steelblue31119180, semithick]
(axis cs:32,0.985203810757357)
--(axis cs:32,0.99646094957687);

\path [draw=steelblue31119180, semithick]
(axis cs:33,0.982903566938507)
--(axis cs:33,0.996339516384543);

\path [draw=steelblue31119180, semithick]
(axis cs:34,0.991963665128518)
--(axis cs:34,0.999466170033948);

\path [draw=steelblue31119180, semithick]
(axis cs:35,0.995504248179451)
--(axis cs:35,0.998921734250077);

\path [draw=steelblue31119180, semithick]
(axis cs:36,0.999966965672948)
--(axis cs:36,1.00000026097981);

\addplot [semithick, steelblue31119180, mark=-, mark size=2, mark options={solid}, only marks]
table {%
1 0.999862696254946
2 -1.17940646205767e-06
3 -0.000444955094551462
4 0.149880880175691
5 0.0148539247429455
6 0.499177556359257
7 0.452171312470224
8 -1.66601383678422e-06
9 -7.69951486654214e-07
10 -3.65594625158578e-07
11 -5.38016069689417e-07
12 -6.50100828623347e-07
13 -8.93660863582813e-07
14 0.0755359300492911
15 0.110353076156504
16 0.317207567538609
17 0.287292123846153
18 0.0103364909681578
19 0.61323344991875
20 0.70748251841905
21 0.545653313424864
22 0.74485186318899
23 0.786037264551979
24 0.846707330917601
25 0.908620153237746
26 0.939624328116863
27 0.958265356894392
28 0.992443957787352
29 0.981409930527113
30 0.974249181864862
31 0.974968447471581
32 0.985203810757357
33 0.982903566938507
34 0.991963665128518
35 0.995504248179451
36 0.999966965672948
};
\addplot [semithick, steelblue31119180, mark=-, mark size=2, mark options={solid}, only marks]
table {%
1 1.0000037370574
2 1.14789987263998e-05
3 0.00064913240596353
4 0.345364907950286
5 0.07314457421611
6 0.645458883771822
7 0.625949293071374
8 3.38226304607863e-05
9 7.45022603211534e-06
10 7.71478202587656e-06
11 1.00360531289019e-05
12 1.26173870491808e-05
13 1.24753944862336e-05
14 0.10331609226345
15 0.155697896417795
16 0.362165645160042
17 0.357888287600546
18 0.0191130795322117
19 0.660760820373833
20 0.754072902629127
21 0.633417729318876
22 0.800586261207122
23 0.848336852074331
24 0.910770201686911
25 0.952636265610178
26 0.972131526360618
27 0.985702041537859
28 0.998326739603532
29 0.993035794414477
30 0.993189633455799
31 0.992635409155749
32 0.99646094957687
33 0.996339516384543
34 0.999466170033948
35 0.998921734250077
36 1.00000026097981
};
\addplot [semithick, steelblue31119180, mark=*, mark size=1, mark options={solid}, only marks]
table {%
1 0.999933216656175
2 5.14979613217105e-06
3 0.000102088655706034
4 0.247622894062989
5 0.0439992494795277
6 0.57231822006554
7 0.539060302770799
8 1.6078308312001e-05
9 3.34013727273056e-06
10 3.67459370035899e-06
11 4.74901852960625e-06
12 5.98364311027873e-06
13 5.79086681132541e-06
14 0.0894260111563706
15 0.133025486287149
16 0.339686606349325
17 0.322590205723349
18 0.0147247852501848
19 0.636997135146291
20 0.730777710524089
21 0.58953552137187
22 0.772719062198056
23 0.817187058313155
24 0.878738766302256
25 0.930628209423962
26 0.955877927238741
27 0.971983699216125
28 0.995385348695442
29 0.987222862470795
30 0.983719407660331
31 0.983801928313665
32 0.990832380167113
33 0.989621541661525
34 0.995714917581233
35 0.997212991214764
36 0.99998361332638
};
\end{axis}

\end{tikzpicture}
        }
        \vskip -0.12in
        \caption{FFN gate probabilities.}
         \label{fig:ffn-gate-prob}
     \end{subfigure}
     \vskip -0.15in
        \caption{Predicted gate probabilities (mean and std) at different layers of an I3D-GlobalGP model on LibriSpeech test other. A higher probability means the layer is more likely to be executed.}
        \label{fig:gate-probs}
        \vskip -0.15in
\end{figure}
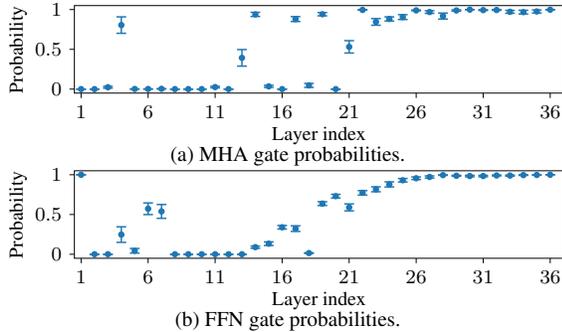

\begin{figure}[t]
     \centering
     \begin{subfigure}[b]{\linewidth}
         \centering
         \resizebox {0.9\linewidth} {!} {
         \begin{tikzpicture}

\definecolor{darkgray176}{RGB}{176,176,176}
\definecolor{steelblue31119180}{RGB}{31,119,180}

\begin{axis}[
tick align=outside,
tick pos=left,
x grid style={darkgray176},
xlabel={Layer index},
xmin=0.5, xmax=37,
xtick style={color=black},
xtick={1,6,11,...,36},
y grid style={darkgray176},
ylabel={Probability},
ymin=-0.0500431648545307, ymax=1.04980744365575,
ytick style={color=black},
major tick length=2pt,
width=\linewidth,
height=0.33\linewidth,
ylabel shift = -3pt,
xlabel shift = -3pt,
label style={font=\footnotesize},
ticklabel style={font=\footnotesize},
]
\path [draw=steelblue31119180, semithick]
(axis cs:1,0.807766544290467)
--(axis cs:1,0.872360805782403);

\path [draw=steelblue31119180, semithick]
(axis cs:2,0.101784537214732)
--(axis cs:2,0.164367871356282);

\path [draw=steelblue31119180, semithick]
(axis cs:3,0.663115534377694)
--(axis cs:3,0.799778075478518);

\path [draw=steelblue31119180, semithick]
(axis cs:4,0.108459100825529)
--(axis cs:4,0.196852446800343);

\path [draw=steelblue31119180, semithick]
(axis cs:5,0.989299905752316)
--(axis cs:5,0.995243268093799);

\path [draw=steelblue31119180, semithick]
(axis cs:6,0.990425908337105)
--(axis cs:6,0.996012748578869);

\path [draw=steelblue31119180, semithick]
(axis cs:7,0.996240755703915)
--(axis cs:7,0.998827234458945);

\path [draw=steelblue31119180, semithick]
(axis cs:8,0.993930694285207)
--(axis cs:8,0.997573996928964);

\path [draw=steelblue31119180, semithick]
(axis cs:9,0.998568139740193)
--(axis cs:9,0.999573565418141);

\path [draw=steelblue31119180, semithick]
(axis cs:10,0.999667699082549)
--(axis cs:10,0.999941909071189);

\path [draw=steelblue31119180, semithick]
(axis cs:11,0.999013702167857)
--(axis cs:11,0.999824769768528);

\path [draw=steelblue31119180, semithick]
(axis cs:12,0.999775537547503)
--(axis cs:12,0.999986419145375);

\path [draw=steelblue31119180, semithick]
(axis cs:13,0.000893952749085521)
--(axis cs:13,0.00314674833174207);

\path [draw=steelblue31119180, semithick]
(axis cs:14,0.0230697998707707)
--(axis cs:14,0.0595357621261887);

\path [draw=steelblue31119180, semithick]
(axis cs:15,0.00318804976484593)
--(axis cs:15,0.00843104782970103);

\path [draw=steelblue31119180, semithick]
(axis cs:16,0.733885507576147)
--(axis cs:16,0.794986222682829);

\path [draw=steelblue31119180, semithick]
(axis cs:17,0.0450839340273302)
--(axis cs:17,0.0815387728765972);

\path [draw=steelblue31119180, semithick]
(axis cs:18,0.817408503526921)
--(axis cs:18,0.884938276490103);

\path [draw=steelblue31119180, semithick]
(axis cs:19,0.840084106057572)
--(axis cs:19,0.880817879993496);

\path [draw=steelblue31119180, semithick]
(axis cs:20,0.771067448738567)
--(axis cs:20,0.819191041858592);

\path [draw=steelblue31119180, semithick]
(axis cs:21,0.858483230401101)
--(axis cs:21,0.914498651694236);

\path [draw=steelblue31119180, semithick]
(axis cs:22,0.92488828733634)
--(axis cs:22,0.958239051617253);

\path [draw=steelblue31119180, semithick]
(axis cs:23,0.831305646788897)
--(axis cs:23,0.903239874900197);

\path [draw=steelblue31119180, semithick]
(axis cs:24,0.927918989243587)
--(axis cs:24,0.964629899546327);

\path [draw=steelblue31119180, semithick]
(axis cs:25,3.65205033114093e-05)
--(axis cs:25,0.000232260578238162);

\path [draw=steelblue31119180, semithick]
(axis cs:26,1.93860945143828e-05)
--(axis cs:26,0.000109549997817988);

\path [draw=steelblue31119180, semithick]
(axis cs:27,6.20671507153683e-06)
--(axis cs:27,9.6340184462525e-05);

\path [draw=steelblue31119180, semithick]
(axis cs:28,3.09855259843665e-05)
--(axis cs:28,0.000154776196630011);

\path [draw=steelblue31119180, semithick]
(axis cs:29,5.88405433072645e-06)
--(axis cs:29,0.000118024214485406);

\path [draw=steelblue31119180, semithick]
(axis cs:30,1.91004571857145e-05)
--(axis cs:30,0.000158900378333247);

\path [draw=steelblue31119180, semithick]
(axis cs:31,1.02219386031434e-05)
--(axis cs:31,0.000124933818151526);

\path [draw=steelblue31119180, semithick]
(axis cs:32,5.36434040145644e-05)
--(axis cs:32,0.000257318120569174);

\path [draw=steelblue31119180, semithick]
(axis cs:33,2.22051845632149e-05)
--(axis cs:33,0.000136830026130426);

\path [draw=steelblue31119180, semithick]
(axis cs:34,3.88469267985369e-06)
--(axis cs:34,7.35033281335822e-05);

\path [draw=steelblue31119180, semithick]
(axis cs:35,1.09450153425813e-06)
--(axis cs:35,4.72503467758096e-05);

\path [draw=steelblue31119180, semithick]
(axis cs:36,1.54215218763314e-05)
--(axis cs:36,0.000154304797298429);

\addplot [semithick, steelblue31119180, mark=-, mark size=2, mark options={solid}, only marks]
table {%
1 0.807766544290467
2 0.101784537214732
3 0.663115534377694
4 0.108459100825529
5 0.989299905752316
6 0.990425908337105
7 0.996240755703915
8 0.993930694285207
9 0.998568139740193
10 0.999667699082549
11 0.999013702167857
12 0.999775537547503
13 0.000893952749085521
14 0.0230697998707707
15 0.00318804976484593
16 0.733885507576147
17 0.0450839340273302
18 0.817408503526921
19 0.840084106057572
20 0.771067448738567
21 0.858483230401101
22 0.92488828733634
23 0.831305646788897
24 0.927918989243587
25 3.65205033114093e-05
26 1.93860945143828e-05
27 6.20671507153683e-06
28 3.09855259843665e-05
29 5.88405433072645e-06
30 1.91004571857145e-05
31 1.02219386031434e-05
32 5.36434040145644e-05
33 2.22051845632149e-05
34 3.88469267985369e-06
35 1.09450153425813e-06
36 1.54215218763314e-05
};
\addplot [semithick, steelblue31119180, mark=-, mark size=2, mark options={solid}, only marks]
table {%
1 0.872360805782403
2 0.164367871356282
3 0.799778075478518
4 0.196852446800343
5 0.995243268093799
6 0.996012748578869
7 0.998827234458945
8 0.997573996928964
9 0.999573565418141
10 0.999941909071189
11 0.999824769768528
12 0.999986419145375
13 0.00314674833174207
14 0.0595357621261887
15 0.00843104782970103
16 0.794986222682829
17 0.0815387728765972
18 0.884938276490103
19 0.880817879993496
20 0.819191041858592
21 0.914498651694236
22 0.958239051617253
23 0.903239874900197
24 0.964629899546327
25 0.000232260578238162
26 0.000109549997817988
27 9.6340184462525e-05
28 0.000154776196630011
29 0.000118024214485406
30 0.000158900378333247
31 0.000124933818151526
32 0.000257318120569174
33 0.000136830026130426
34 7.35033281335822e-05
35 4.72503467758096e-05
36 0.000154304797298429
};
\addplot [semithick, steelblue31119180, mark=*, mark size=1, mark options={solid}, only marks]
table {%
1 0.840063675036435
2 0.133076204285507
3 0.731446804928106
4 0.152655773812936
5 0.992271586923057
6 0.993219328457987
7 0.99753399508143
8 0.995752345607085
9 0.999070852579167
10 0.999804804076869
11 0.999419235968192
12 0.999880978346439
13 0.00202035054041379
14 0.0413027809984797
15 0.00580954879727348
16 0.764435865129488
17 0.0633113534519637
18 0.851173390008512
19 0.860450993025534
20 0.79512924529858
21 0.886490941047668
22 0.941563669476796
23 0.867272760844547
24 0.946274444394957
25 0.000134390540774785
26 6.44680461661856e-05
27 5.12734497670309e-05
28 9.28808613071886e-05
29 6.19541344080664e-05
30 8.90004177594806e-05
31 6.75778783773346e-05
32 0.000155480762291869
33 7.95176053468202e-05
34 3.86940104067179e-05
35 2.41724241550339e-05
36 8.48631595873801e-05
};
\end{axis}

\end{tikzpicture}
         }
         \vskip -0.12in
         \caption{MHA gate probabilities (trained with InterCTC).}
         \label{fig:mha-gate-prob-interctc}
     \end{subfigure}

     \vskip -0.05in
     \begin{subfigure}[b]{\linewidth}
         \centering
         \resizebox {0.9\linewidth} {!} {
        \begin{tikzpicture}

\definecolor{darkgray176}{RGB}{176,176,176}
\definecolor{steelblue31119180}{RGB}{31,119,180}

\begin{axis}[
tick align=outside,
tick pos=left,
x grid style={darkgray176},
xlabel={Layer index},
xmin=0.5, xmax=37,
xtick style={color=black},
xtick={1,6,11,...,36},
y grid style={darkgray176},
ylabel={Probability},
ymin=-0.0504673897021493, ymax=1.050026171665,
ytick style={color=black},
major tick length=2pt,
width=\linewidth,
height=0.33\linewidth,
ylabel shift = -3pt,
xlabel shift = -3pt,
label style={font=\footnotesize},
ticklabel style={font=\footnotesize},
]
\path [draw=steelblue31119180, semithick]
(axis cs:1,0.997074972939576)
--(axis cs:1,0.999220465549789);

\path [draw=steelblue31119180, semithick]
(axis cs:2,0.649006539476554)
--(axis cs:2,0.80322371782026);

\path [draw=steelblue31119180, semithick]
(axis cs:3,0.705428466412817)
--(axis cs:3,0.794772260695609);

\path [draw=steelblue31119180, semithick]
(axis cs:4,0.525149816541701)
--(axis cs:4,0.644709478631151);

\path [draw=steelblue31119180, semithick]
(axis cs:5,0.875378091693246)
--(axis cs:5,0.924377491963119);

\path [draw=steelblue31119180, semithick]
(axis cs:6,0.943285372510396)
--(axis cs:6,0.96700316402605);

\path [draw=steelblue31119180, semithick]
(axis cs:7,0.959585120569051)
--(axis cs:7,0.981472251378129);

\path [draw=steelblue31119180, semithick]
(axis cs:8,0.98876674093946)
--(axis cs:8,0.996186495705332);

\path [draw=steelblue31119180, semithick]
(axis cs:9,0.997715492589142)
--(axis cs:9,0.999273958655701);

\path [draw=steelblue31119180, semithick]
(axis cs:10,0.998956734128981)
--(axis cs:10,0.999823433797224);

\path [draw=steelblue31119180, semithick]
(axis cs:11,0.999198011117326)
--(axis cs:11,0.999805054573737);

\path [draw=steelblue31119180, semithick]
(axis cs:12,0.999492815827717)
--(axis cs:12,0.999960158661124);

\path [draw=steelblue31119180, semithick]
(axis cs:13,0.30932756195379)
--(axis cs:13,0.420443443581627);

\path [draw=steelblue31119180, semithick]
(axis cs:14,0.590157576271691)
--(axis cs:14,0.67723367855394);

\path [draw=steelblue31119180, semithick]
(axis cs:15,0.764347627250401)
--(axis cs:15,0.810648539986386);

\path [draw=steelblue31119180, semithick]
(axis cs:16,0.482875226119936)
--(axis cs:16,0.604398081260468);

\path [draw=steelblue31119180, semithick]
(axis cs:17,0.678203382027576)
--(axis cs:17,0.779750037857349);

\path [draw=steelblue31119180, semithick]
(axis cs:18,0.689403823645046)
--(axis cs:18,0.760761741209028);

\path [draw=steelblue31119180, semithick]
(axis cs:19,0.69891554666666)
--(axis cs:19,0.779142387468772);

\path [draw=steelblue31119180, semithick]
(axis cs:20,0.837131538592611)
--(axis cs:20,0.885340366267486);

\path [draw=steelblue31119180, semithick]
(axis cs:21,0.826291549768021)
--(axis cs:21,0.893233510093912);

\path [draw=steelblue31119180, semithick]
(axis cs:22,0.9446087255317)
--(axis cs:22,0.965781827021177);

\path [draw=steelblue31119180, semithick]
(axis cs:23,0.919442506683367)
--(axis cs:23,0.95962369383095);

\path [draw=steelblue31119180, semithick]
(axis cs:24,0.998184704805175)
--(axis cs:24,0.999540413816983);

\path [draw=steelblue31119180, semithick]
(axis cs:25,5.51773137444499e-07)
--(axis cs:25,5.37764567837133e-05);

\path [draw=steelblue31119180, semithick]
(axis cs:26,2.98862504644789e-06)
--(axis cs:26,6.42613271024909e-05);

\path [draw=steelblue31119180, semithick]
(axis cs:27,1.23944705729061e-05)
--(axis cs:27,0.000133745101365698);

\path [draw=steelblue31119180, semithick]
(axis cs:28,1.11914350527547e-05)
--(axis cs:28,0.000180805452011943);

\path [draw=steelblue31119180, semithick]
(axis cs:29,-2.58129973674871e-06)
--(axis cs:29,7.26530637254616e-05);

\path [draw=steelblue31119180, semithick]
(axis cs:30,6.85065039644912e-07)
--(axis cs:30,5.83244665774764e-05);

\path [draw=steelblue31119180, semithick]
(axis cs:31,-2.54748943081287e-06)
--(axis cs:31,2.42960324851761e-05);

\path [draw=steelblue31119180, semithick]
(axis cs:32,9.06240828239411e-06)
--(axis cs:32,9.8239924698761e-05);

\path [draw=steelblue31119180, semithick]
(axis cs:33,7.00882209847724e-06)
--(axis cs:33,9.65303802477634e-05);

\path [draw=steelblue31119180, semithick]
(axis cs:34,-1.44579752503401e-07)
--(axis cs:34,0.000139044010887156);

\path [draw=steelblue31119180, semithick]
(axis cs:35,2.38592849303532e-06)
--(axis cs:35,7.05216111533491e-05);

\path [draw=steelblue31119180, semithick]
(axis cs:36,3.0239548830191e-05)
--(axis cs:36,0.000224675926436248);

\addplot [semithick, steelblue31119180, mark=-, mark size=2, mark options={solid}, only marks]
table {%
1 0.997074972939576
2 0.649006539476554
3 0.705428466412817
4 0.525149816541701
5 0.875378091693246
6 0.943285372510396
7 0.959585120569051
8 0.98876674093946
9 0.997715492589142
10 0.998956734128981
11 0.999198011117326
12 0.999492815827717
13 0.30932756195379
14 0.590157576271691
15 0.764347627250401
16 0.482875226119936
17 0.678203382027576
18 0.689403823645046
19 0.69891554666666
20 0.837131538592611
21 0.826291549768021
22 0.9446087255317
23 0.919442506683367
24 0.998184704805175
25 5.51773137444499e-07
26 2.98862504644789e-06
27 1.23944705729061e-05
28 1.11914350527547e-05
29 -2.58129973674871e-06
30 6.85065039644912e-07
31 -2.54748943081287e-06
32 9.06240828239411e-06
33 7.00882209847724e-06
34 -1.44579752503401e-07
35 2.38592849303532e-06
36 3.0239548830191e-05
};
\addplot [semithick, steelblue31119180, mark=-, mark size=2, mark options={solid}, only marks]
table {%
1 0.999220465549789
2 0.80322371782026
3 0.794772260695609
4 0.644709478631151
5 0.924377491963119
6 0.96700316402605
7 0.981472251378129
8 0.996186495705332
9 0.999273958655701
10 0.999823433797224
11 0.999805054573737
12 0.999960158661124
13 0.420443443581627
14 0.67723367855394
15 0.810648539986386
16 0.604398081260468
17 0.779750037857349
18 0.760761741209028
19 0.779142387468772
20 0.885340366267486
21 0.893233510093912
22 0.965781827021177
23 0.95962369383095
24 0.999540413816983
25 5.37764567837133e-05
26 6.42613271024909e-05
27 0.000133745101365698
28 0.000180805452011943
29 7.26530637254616e-05
30 5.83244665774764e-05
31 2.42960324851761e-05
32 9.8239924698761e-05
33 9.65303802477634e-05
34 0.000139044010887156
35 7.05216111533491e-05
36 0.000224675926436248
};
\addplot [semithick, steelblue31119180, mark=*, mark size=1, mark options={solid}, only marks]
table {%
1 0.998147719244683
2 0.726115128648407
3 0.750100363554213
4 0.584929647586426
5 0.899877791828183
6 0.955144268268223
7 0.97052868597359
8 0.992476618322396
9 0.998494725622422
10 0.999390083963103
11 0.999501532845531
12 0.999726487244421
13 0.364885502767709
14 0.633695627412815
15 0.787498083618394
16 0.543636653690202
17 0.728976709942462
18 0.725082782427037
19 0.739028967067716
20 0.861235952430049
21 0.859762529930966
22 0.955195276276438
23 0.939533100257158
24 0.998862559311079
25 2.71641149605789e-05
26 3.36249760744694e-05
27 7.3069785969302e-05
28 9.59984435323486e-05
29 3.50358819943565e-05
30 2.95047658085606e-05
31 1.08742715271816e-05
32 5.36511664905775e-05
33 5.17696011731203e-05
34 6.94497155673261e-05
35 3.64537698231922e-05
36 0.00012745773763322
};
\end{axis}

\end{tikzpicture}
        }
        \vskip -0.12in
        \caption{FFN gate probabilities (trained with InterCTC).}
         \label{fig:ffn-gate-prob-interctc}
     \end{subfigure}
     \vskip -0.15in
        \caption{Predicted gate probabilities (mean and std) at different layers of an I3D-GlobalGP model with InterCTC on LibriSpeech test other. A higher probability means the layer is more likely to be executed.}
        \label{fig:gate-probs-interctc}
        \vskip -0.15in
\end{figure}
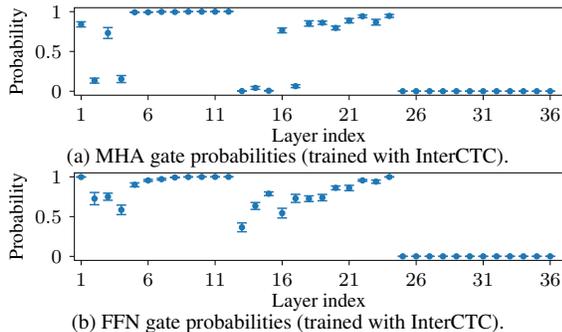

\begin{figure}[t]
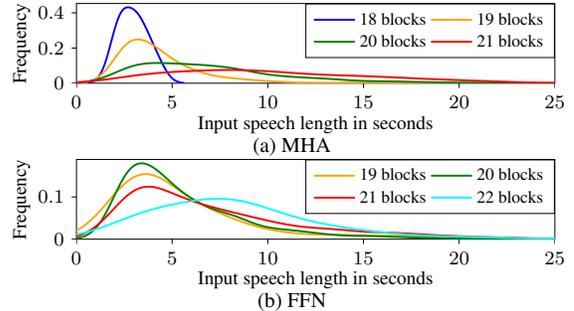

     \centering
     \begin{subfigure}[b]{\linewidth}
         \centering
         \resizebox {0.9\linewidth} {!} {
         \input{figures/lengthdensity_att.tex}
         }
         \vskip -0.11in
        \caption{MHA}
        \label{fig:length-mha}
     \end{subfigure}

     \vskip -0.05in
     \begin{subfigure}[b]{\linewidth}
         \centering
         \resizebox {0.9\linewidth} {!} {
         \input{figures/lengthdensity_ffn.tex}
         }
          \vskip -0.11in
         \caption{FFN}
         \label{fig:length-ffn}
     \end{subfigure}
     \vskip -0.15in
        \caption{Distributions of speech lengths categorized by the number of MHA or FFN blocks used for inference. This is an I3D-GlobalGP model evaluated on LibriSpeech test other. Utterances using more blocks tend to be longer.}
        \label{fig:length-analysis}
        \vskip -0.15in
\end{figure}

\subsection{Experimental setup}

We use PyTorch~\cite{pytorch} and follow the ASR recipes in ESPnet~\cite{espnet} to train all models. We mainly use the CTC framework on LibriSpeech 100h~\cite{librispeech-corpus}. In Sec.~\ref{subsec:generalizability}, we also show that I3D can be applied to AED and another corpus, Tedlium2~\cite{tedlium2}.
Our I3D encoders have 36 layers in total. They are initialized with trained standard Transformers and fine-tuned with a reduced learning rate ($1e-3$) and various $\lambda$ (usually ranging from 1 to 13) in Eq.~\eqref{eq:total-loss} to trade off WER and computation. The fine-tuning epochs for LibriSpeech 100h and Tedlium2 are 50 and 35, respectively.
We compare I3D with two baselines.
First, we train standard Transformers with a reduced number of layers.
Second, we train a 36-layer Transformer with LayerDrop~\cite{huang2016deep,fan2020Reducing} or Intermediate CTC (InterCTC)~\cite{interctc} and perform iterative layer pruning~\cite{lee2021layer} using the validation set to get a variety of models with smaller and fixed architectures.
This baseline is denoted as ``LayerDrop'' in Figs.~\ref{fig:ls-ctc} and~\ref{fig:ls-interctc}.
We can compare I3D, whose layers are dynamically reduced based on the input, against the static pruned models.

\subsection{Main results}
\label{subsec:main-results}

Fig.~\ref{fig:ls-ctc} compares our I3D models with two baselines. We train I3D-CTC models with different $\lambda$ in Eq.~\eqref{eq:total-loss} to adjust the operating point. We calculate the number of layers as the average of the number of MHA blocks and the number of FFN blocks.
Both I3D-LocalGP and I3D-GlobalGP outperform the standard Transformer and the pruned version using iterative layer pruning~\cite{lee2021layer}. We can reduce the average number of layers to around 20 while still matching the Transformer trained from scratch. LocalGP achieves similar performance as GlobalGP, but GlobalGP has only one gate predictor, which can be more efficient for inference.
The reason why LocalGP is not better than GlobalGP may be that LocalGP decides whether to execute or skip a block based on the current layer's input, which depends on decisions at previous layers. This sequential procedure can lead to more severe error propagation.
We also show that it is possible to adjust the computational cost of a trained I3D model by changing $\beta$ (see Eqs.~\eqref{eq:mha-threshold}~\eqref{eq:ffn-threshold}) at inference time. Three I3D-GlobalGP models are decoded with different $\beta$. As $\beta$ decreases, more blocks are used, and the WER is usually improved.

Fig.~\ref{fig:ls-interctc} shows the results using InterCTC~\cite{interctc}. The WERs are lower than those in Fig.~\ref{fig:ls-ctc}, thanks to the auxiliary CTC loss which regularizes training. Again, I3D is consistently better than the Transformer trained from scratch and the pruned model.

\subsection{Analysis of gate distributions}
\label{subsec:gate-distributions}

Fig.~\ref{fig:gate-probs} shows the mean and standard deviation (std) of the gate probabilities generated by an I3D-GlobalGP model using CTC on LibriSpeech test-other. Most layers have a stable probability. Several layers have larger variations depending on the input.
For both MHA and FFN, upper layers are executed with high probabilities while lower layers tend to be skipped, which is consistent with~\cite{xie22interspeech-streaming}.

We also show the gate probabilities from an I3D-GlobalGP model trained with InterCTC~\cite{interctc} in Fig.~\ref{fig:gate-probs-interctc}. Interestingly, the overall trend is very different from Fig.~\ref{fig:gate-probs}. Now, the upper layers are almost skipped while the lower layers are executed with very high probabilities, indicating that lower layers of this encoder can learn powerful representations for the ASR task. This is probably because auxiliary CTC losses inserted at intermediate layers can facilitate the gradient propagation to lower parts of a deep encoder, which effectively improves its capacity and also the final performance.

We believe this gate analysis can provide a way to interpret the layer-wise behavior of deep networks.

\subsection{Analysis of input-dependency}
\label{subsec:analysis-input-dependency}

It has been shown that our I3D models can dynamically adjust the encoder depth based on the characteristics of an input utterance, which achieves strong performance even with reduced computation. But it is unclear which features are important for the gate predictor to determine the modules used during inference. 
We have found that the speech length generally affects the inference architecture. Fig.~\ref{fig:length-analysis} shows the speech length distributions categorized by the number of MHA or FFN blocks used by an I3D-GlobalGP model during inference. We observe that utterances using more blocks tend to be longer. This is probably because longer utterances contain more complex information and longer-range dependency among frames, which require more blocks (especially MHA) to process.

We also considered two other factors that may affect the inference architecture, namely the difficulty of utterances measured by WERs, and the audio quality measured by DNSMOS scores~\cite{dnsmos}. However, in general, we didn't observe a clear relationship between these metrics and the number of layers used for inference.

\subsection{Generalizability}
\label{subsec:generalizability}

We demonstrate that the proposed I3D encoders can be directly applied to other datasets and ASR frameworks. Fig.~\ref{fig:tedlium-ctc-test} shows the results of CTC-based models on Tedlium2. Our I3D models consistently achieve lower WERs than the standard Transformer with similar or even fewer layers during inference.~\footnote{We have also evaluated I3D on LibriSpeech 960h. Observations are consistent with LibriSpeech 100h and Tedlium2.} We further apply I3D to the attention-based encoder-decoder (AED) framework. Only the encoder is changed while the decoder is still a standard Transformer decoder. Table~\ref{tab:ls100-aed} presents the results on LibriSpeech 100h. With around 27 layers on average during inference, our I3D models outperform the 27-layer Transformer trained from scratch on both dev clean and test clean sets. The I3D with a global gate predictor is slightly better than that with a local gate predictor.

\section{Conclusion}
\label{sec:conclusion}

In this work, we propose I3D, a Transformer-based encoder which dynamically adjusts its depth based on the characteristics of input utterances to trade off performance and efficiency. We design two types of gate predictors and show that I3D-based models consistently outperform the vanilla Transformer trained from scratch and the static pruned model. I3D can be applied to various end-to-end ASR frameworks and corpora. We also present interesting analysis on the predicted gate probabilities and the input-dependency to better interpret the behavior of deep encoders and the effect of intermediate loss regularization techniques. In the future, we plan to apply this method to large pre-trained models. We will explore only fine-tuning gate predictors to significantly reduce training cost.

\section{Acknowledgements}

This work used Bridges2 at PSC and Delta at NCSA through allocation CIS210014 from the Advanced Cyberinfrastructure Coordination Ecosystem: Services \& Support (ACCESS) program, which is supported by National Science Foundation grants \#2138259, \#2138286, \#2138307, \#2137603, and \#2138296.

\newpage

\bibliographystyle{IEEEbib}
\bibliography{refs}

\end{document}